\pdfoutput=1
\documentclass[11pt]{article}

\usepackage[preprint]{acl}

\usepackage{times}
\usepackage{latexsym}
\usepackage{multirow}
\usepackage{booktabs}
\usepackage{tabularx}

\usepackage[T1]{fontenc}
\usepackage[utf8]{inputenc}

\usepackage{microtype}
\usepackage{graphicx}

\usepackage{inconsolata}

\usepackage{algorithm}
\usepackage{algpseudocode}
\usepackage{amsmath}
\usepackage{graphicx}
\usepackage{subcaption} % or use subfig

\title{AS-ES Learning: Towards Efficient CoT Learning in Small Models}

\author{
Nuwa Xi\thanks{ \ \ Equal Contribution}, Yuhan Chen\footnotemark[1], Sendong Zhao\thanks{ \ \ Corresponding author}, Haochun Wang, Bing Qin and Ting Liu
 \\
Research Center for Social Computing and Information Retrieval, \\Harbin Institute of Technology, China 
\\
\texttt{\{nwxi,yuhanchen,sdzhao,hcwang,bqin,tliu\}@ir.hit.edu.cn}
}
        
\begin{document}
\maketitle
\begin{abstract}
    Chain-of-Thought (CoT) serves as a critical emerging ability in LLMs, especially when it comes to logical reasoning. Attempts have been made to induce such ability in small models as well by distilling from the data with CoT generated by Large Language Models (LLMs). However, existing methods often simply generate and incorporate more data from LLMs and fail to note the importance of efficiently utilizing existing CoT data. We here propose a new training paradigm AS-ES (Abstractive Segments - Extractive Segments) learning, which exploits the inherent information in CoT for iterative generation. Experiments show that our methods surpass the direct seq2seq training on CoT-extensive tasks like MWP and PET summarization, without data augmentation or altering the model itself. Furthermore, we explore the reason behind the inefficiency of small models in learning CoT and provide an explanation of why AS-ES learning works, giving insights into the underlying mechanism of CoT.
\end{abstract}
\section{Introduction}
\label{intro}

CoT is one of the most important emerging abilities that distinguishes LLMs from prior models with smaller scales \cite{wei2022chain}. The explicit introduction of CoT enables LLMs to tackle complex problems that necessitate critical thinking and intricate logical reasoning, thus enhancing the overall performance of LLMs. \cite{wei2022chain, Zhang2023Multimodal, Wang2022Self-Consistency}.

Intuitively, numerous works extended the CoT capacity to smaller-scale models \cite{shridhar2023distilling, hsieh2023distilling, fu2023specializing, ma2023sci, chen2023mcc, wang-etal-2022-smash, ho2022large}. Yet, these endeavors predominantly concentrate on generating an increased quantity of high-quality CoT data from LLMs and resort to a direct seq2seq training approach, where the query forms the input and the CoT-enriched answer serves as the target. Such methods often overlook the limited capacity of small models in learning the complex reasoning in CoT. Taking this into account, Some works tried to decompose CoT into more fine-grained reasoning steps, and used LLMs to generate rationales for each individual step to augment the original CoT \cite{hsieh2023distilling, ma2023sci, Zhang2023Multimodal}. \citet{shridhar2023distilling} further improved the direct seq2seq paradigm using the augmented CoT to train separate models for iterative generation. However, the use of augmented CoT is just another way of generating more CoT data, which is costly and still fails to fully exploit the inherent information in existing datasets. 

Furthermore, although proven to achieve better performance, using separate small models for CoT learning raises another question. Similar to using different modules for information retrieval and reasoning in multi-hop QA \cite{deng2020multi, jiang2019self, feng-etal-2020-scalable,mavi2022survey}, these methods specialize small models in single-task operations, in contrast with LLMs using a singular framework. Such implementation underlines an assumption that small models, due to their constrained computational capacity compared to LLMs, may require separate, specialized models to perform distinct phases of CoT processing \cite{Weichert2019A, fu2023specializing}.

% While LLMs seem adept at both extraction and reasoning within just one model, small models typically  \cite{Schick2020It’s, Weichert2019A}, such as . This aligns the methods of using two models for iterative generation of each reasoning step in CoT. 

\begin{figure*}[t]
  \centering
  \includegraphics[width=\linewidth]{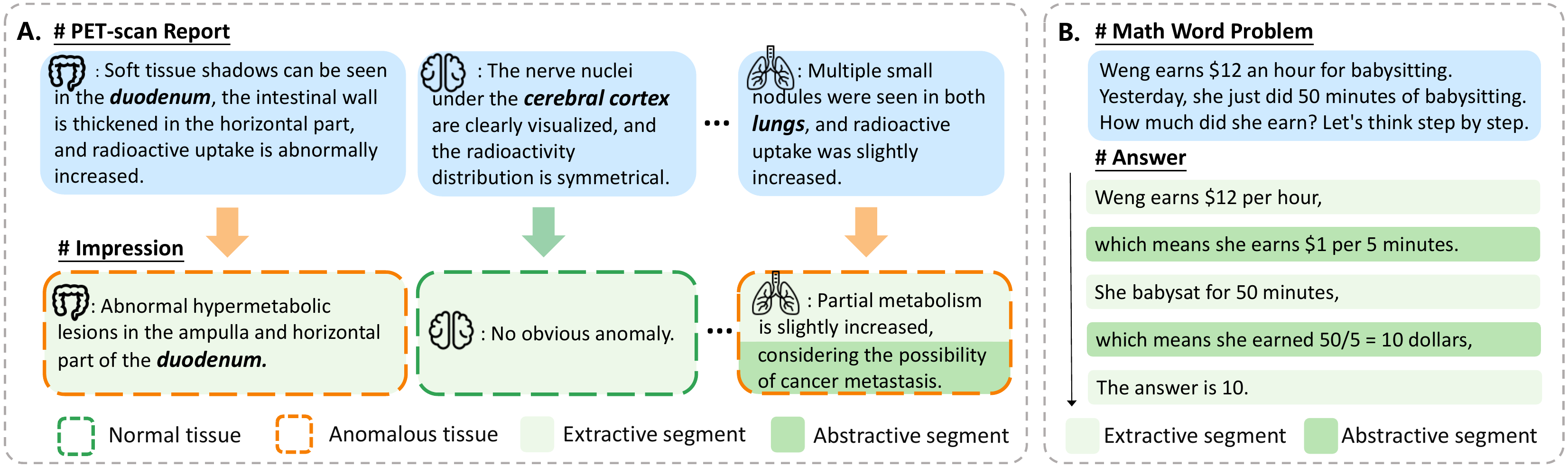}
  \caption{
  \textbf{A.} Description of PET-scan summarization task: Each blue section denotes the part of the report involving a particualr organ, with the green representing the related impression. Organ names are \textbf{\textit{bolded and italicized}}, while light and dark green distinguish between impression segments. \textbf{B.} Description of Math Word Problem task: The blue section highlights the question, while light and dark green sections denote distinct segments of the answer.
  }
  \label{fig:task_description}
\end{figure*}
In response to the above issues, we introduce a novel paradigm different from earlier approaches for distilling CoT to small models. We classify the statements inside the CoT-format results of LLMs into two categories: Extractive Segments (ES) that remind the model of the context and set the stage for subsequent conclusions, and Abstractive Segments (AS) that infer additional insights not explicitly stated in the context. With the deconstruction of CoT into AS and ES components, we curate a dataset tailored for an iterative learning process, defined as AS-ES learning to maximize the latent potential of small models for CoT-intensive tasks without the need for additional data. 

We further experiment with two training strategies, using two models for generating AS/ES respectively and using one unified model for generating AS/ES together, as an attempt to answer whether a single small model can handle both extraction and abstractive reasoning effectively akin to LLMs, and furthermore, whether the suboptimal performance of small models trained with CoT stems from intrinsic limitations or data utilization inefficiencies.

To cover different scenarios, we take two representative problems -- impression generation for PET scan report (PET) and Math Word Problem (MWP). As shown in Figure \ref{fig:task_description}, the CoT of the two tasks expands in different ways: MWP requires a step-by-step solution with CoT following a sequential pattern, while PET involves more parallel processing across multiple body regions.

% The latter can range from formulating recommendations based on detected anomalies in PET scans to performing calculations and logical deductions in MWP, as shown in Fig \ref{fig:task_description}.

In summary, the main contributions of our paper are as follows:

\begin{itemize}
    \item We introduce AS-ES learning, a novel data-efficient training paradigm that maximizes the intrinsic value of existing CoT data, adaptable across various model sizes and tasks.
    \item We explore the use of AS-ES dataset and find that the limitations in CoT learning previously attributed to the inherent capabilities of small models can be substantially mitigated through an improved data utilization strategy without additional data.
    \item We provide a theoretical foundation for the efficacy of AS-ES learning, offering insights into the underlying dynamics of CoT that may benefit future research in the field.
\end{itemize}

\section{Related Work}
\paragraph{Deduction in NLP Tasks}
Deduction is a logical process of reasoning or inferring specific information from given premises or data. In NLP, deduction is used in various tasks, such as summarization, question answering and information extraction \cite{mirzaee2023disentangling, minervini2020learning, deng2020multi, mavi2022survey, qu2020rnnlogic, nye2021improving}. Early research employs end-to-end models to directly learn reasoning strategies from labeled and structured data~\cite{minervini2020learning, qu2020rnnlogic}, which requires the effort of human annotation. \citet{nye2021improving} separates the inference process into a traditional generation part and an extra validation part. They construct a symbolic reasoning module to validate the generated facts using a minimal world model. However, the minimal world model must be hand-engineered. In our work, we employ a fully automatic strategy to disentangle the deduction process, achieving both efficiency and effectiveness.

\paragraph{CoT in Small Models}
CoT is a significant ability to improve the performance of complex reasoning, which is considered as an emergent ability of LLMs \cite{wei2022chain}. Numerous work aims to transfer this ability to small models~\cite{shridhar2023distilling, hsieh2023distilling, fu2023specializing, ma2023sci, chen2023mcc}.
One prevailing method is distilling from LLMs, such as utilizing LLMs to generate rationales or multi-step solutions as training data for smaller models \cite{shridhar2023distilling, hsieh2023distilling}. \citet{chen2023mcc} also allows LLMs to generate multiple rationales whose consistency is enforced by KL-divergence. \citet{ma2023sci} employ a two-stage distillation strategy, in which LLMs not only generate rationale but also provide an answer inference process according to the rationale generated by the student model. However, all these methods require LLMs as a teacher model. In our work, we require no additional data generated by LLMs, which substantially reduces the cost of time and computational resources.

% \citet{shridhar2023distilling} enforce LLMs to produce a multi-step solution, which is used to train smaller models. \citet{hsieh2023distilling} utilize LLMs to generate rationales for given questions and use them as additional training data for small models. Similar strategy is employed and 

% \paragraph{Data Efficiency in Seq-to-seq training}

\section{Method}

\begin{figure*}[t]
  \centering
  \includegraphics[width=\linewidth]{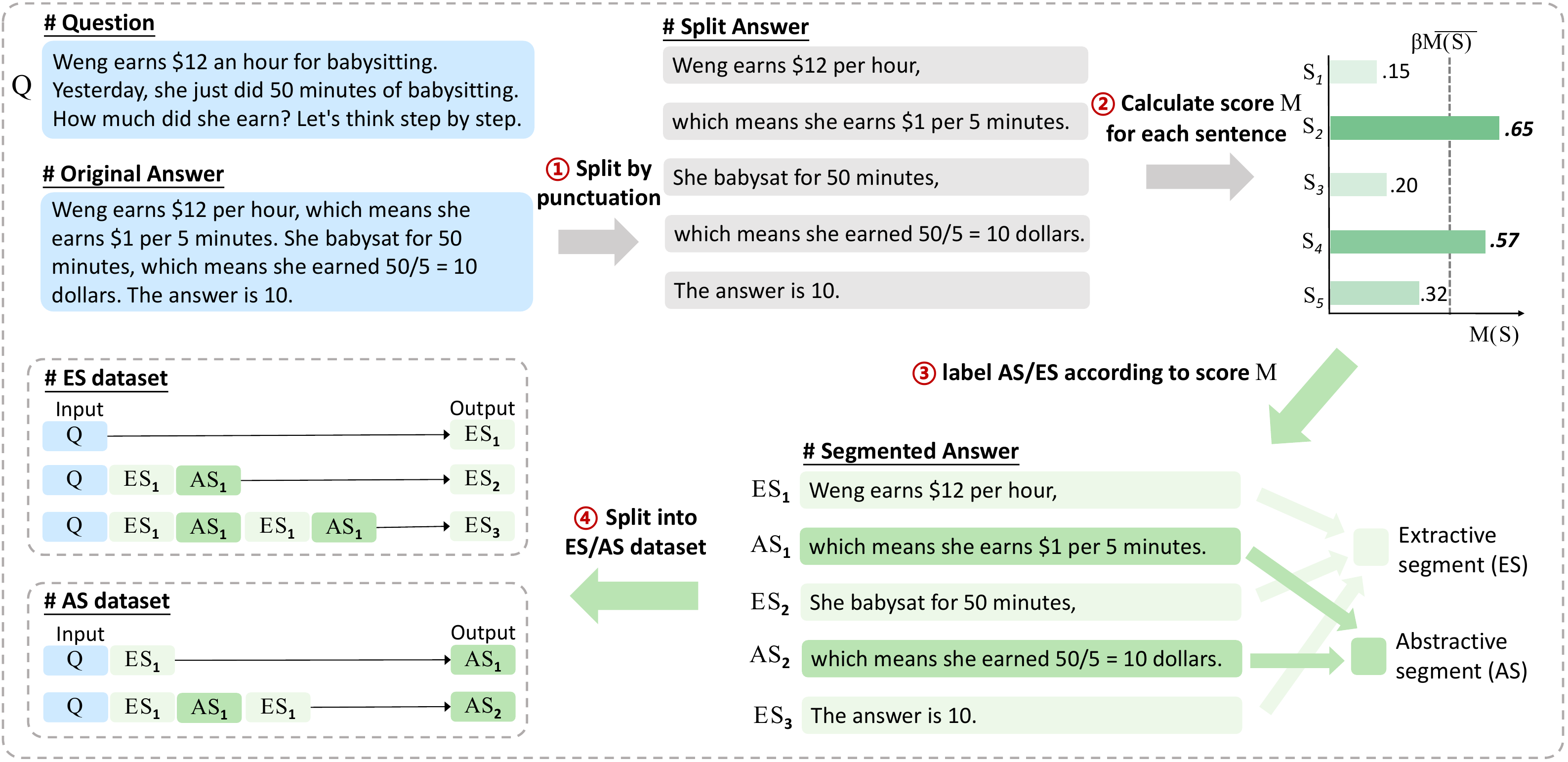}
  \caption{
  The workflow for labeling raw data as either ES or AS parts, followed by constructing ES/AS datasets.
  }
  \label{fig:es_as_split}
\end{figure*}
This section elucidates the methodology of our study, addressing three pivotal questions: (1) How to segment the complete CoT into AS and ES? (2) How to construct a dataset using the segmented AS and ES? (3) How does the AS-ES learning work for training and generation?

\subsection{AS-ES Segmentation}

We explore a variety of segmentation techniques based on distinct characteristics of AS and ES. After splitting all sub-sentences $S_i$ in the complete CoT $S$ by punctuation, we calculate the respective score $M$ using different metrics based on different segmentation methods. Most segmentation methods follow the following criteria ($\beta$ here is a pre-determined hyperparameter).

\begin{equation}
\label{as-equation}
    AS=\{S_i|M(S_i)>\beta \overline{M(S)}\}
\end{equation}

\begin{equation}
\label{es-equation}
    ES=\{S_i|M(S_i)<=\beta \overline{M(S)}\}
\end{equation}

We next introduce the specific segmentation strategies we experiment with. Each strategy is denoted by an abbreviation that will be used subsequently to identify the segmentation applied.

\paragraph{Entropy-oriented Segmentation (ent/ent*)}

Entropy measures uncertainty, and is intrinsically linked to the cross-entropy loss commonly employed in seq2seq training. Denote the input query as $Q$ and the generated response as $R$. We can approximate $P(R|Q)$ by Equation \ref{eq:prob}, and calculate the entropy for the response as in Equation \ref{eq:entropy}.

% \begin{equation}
% \label{eq:loss}
%     l_n=-\log P(R_n|Q)
% \end{equation}

\begin{equation}
\label{eq:prob}
    P(R|Q)\approx softmax(logits(R|Q))
\end{equation}

\begin{equation}
\label{eq:entropy}
    H_n=-\sum_{\forall i \in |V|} P(R_i|Q)\log P(R_i|Q)
\end{equation}

 The rationale is that trained models should exhibit greater certainty regarding the ES, which closely mirrors the input query, thereby resulting in lower entropy compared to AS. This is a way that measures the ``extractive'' and ``abstractive'' from the perspective of a model rather than a human one. The two can diverge largely from each other, and the latter one is almost impossible due to the large cost of human annotation. For comparison, we also employ this segmentation using pre-trained but not fine-tuned models, designated as ent*.

\paragraph{Location-oriented Segmentation (inter)}

Many CoTs naturally fall into an $ES|AS|ES|AS|...$ pattern, especially where in-depth and step-by-step reasoning is required. Each sentence typically presents a complete reasoning step, beginning with context followed by deduction. We exploit this pattern, designating sub-sentences as ES and AS in an interleaving fashion.

\paragraph{Loss-oriented Segmentation (loss)} This approach is akin to entropy-oriented segmentation, where the loss of the trained model is used to estimate its certainty about a sub-sentence. We hypothesize that the efficacy of AS-ES learning may be attributed to its ability to lower the loss boundary, as will be shown in Section \ref{dis:mechanism}. Therefore, segmenting based on loss could potentially enable the model to concentrate more effectively on less well-understood segments.
% suggesting that small models may struggle with concurrent extraction and reasoning.

\paragraph{Similarity-oriented Segmentation (bleu/rouge)}

ES typically derives directly from the original context, sharing greater similarity with the query, whereas AS, often involves new reasoning, making it more different from the query. We utilize BLEU and ROUGE scores to quantify the similarity between the query and the CoT segments.

\subsection{AS-ES Dataset Construction}

\paragraph{Training Data Organization}
Mere segmentation into AS and ES is insufficient for effective model training, as the formats of input and target of the training data play an important role in the training results. After segmenting the original targets into AS and ES, we construct the AS-ES dataset used for AS-ES learning. As shown in Figure \ref{fig:es_as_split}, we first merge the adjacent AS or ES as one, and then structure the data as $Q|ES_1 AS_1 ... ES_{i-1}AS_{i-1}\rightarrow ES_i $ and  $Q|ES_1 AS_1 ... ES_{i-1}AS_{i-1}ES_i\rightarrow AS_i $. Both the ES dataset and AS dataset are compiled from samples generating ES and AS respectively.

\begin{algorithm}
\caption{Dual-Path Generation Process}\label{alg:dual_path}
\begin{algorithmic}[1]
\Require ESM, ASM \Comment{Two models gained from dual-path learning}
\State input $\gets$ Start token
\While{True}
    \State ESM\_output $\gets$ ESM(input)
    \State input $\gets$ input + ESM\_output
    \If{stop\_sign \textbf{in} ESM\_output}
        \State output $\gets$ input
        \Return output
    \EndIf
    \State ASM\_output $\gets$ ASM(input)
    \State input $\gets$ input + ASM\_output
    \If{stop\_sign \textbf{in} ESM\_output}
        \State output $\gets$ input
        \Return output
    \EndIf
\EndWhile
\end{algorithmic}
\end{algorithm}
\paragraph{Stop Sign}

Another question is when to put a stop to iterative generation. We employ a stop sign at the end of sequences generating the final sentence of the original CoT. In this way, the loop can stop once this stop sign is detected during generation. For MWP, the conclusive phrase ``the answer is ...'' serves as this marker. In PET summarization where no inherent stop sign exists, we introduce a special token $\mathtt{<STOP>}$ as the end of each CoT.

\paragraph{Irrelevant Information Processing}

One problem of the AS-ES dataset is that it is rather lengthy compared to the original dataset, which often leads to the problem of exceeding the maximum input length of the model. PET summarization serves as a good example. Due to the limitation of the input length and consideration and the training cost, we first divide whole-body PET scans into sections according to anatomical regions, with each section treated as an independent CoT instance. Unlike MWP, where the entire context provided in the query is integral to formulating the output, PET scan reports typically contain extensive normality, which are generally not included in the final impression. To address this discrepancy, segments depicting normal findings are annotated with ``No obvious anomaly'' as the ground truth, and then incorporated proportionally to ES-dataset. This enables the ES model to identify and selectively extract sentences to the final summary. The proportion of normality incorporated is denoted by $\gamma$, which is the ratio of normal findings included compared to the total number of PET reports.

\subsection{AS-ES Learning}
\paragraph{Dual-path Learning} Existing work that uses iterative generation for CoT learning in small models uses two separate models, one for heuristic questioning and the other for answering. AS-ES training can also adapt to this paradigm by training two separate models, Extractive Segment generation Model (ESM) for retrieving  and Abstractive Segment generation Model (ASM) for reasoning, together designated as Dual-path Segment generation Models (DSM). This training paradigm reflects an underlying assumption that a singular small model may struggle with the complexity of performing both extraction and reasoning tasks within an iterative sequence. The generation process under the DSM framework, as depicted in Algorithm \ref{alg:dual_path}, mirrors the procedural logic of the AS-ES dataset construction. The training set of ASM and ESM can be denoted as follows.

\begin{equation}
    tr(ASM)=\{D_{AS}\}
\end{equation}
\begin{equation}
    tr(ESM)=\{D_{ES}\}
\end{equation}

\paragraph{Uni-path Learning} While the dual-path learning approach has demonstrated efficacy, it necessitates the training and maintenance of two separate models, each with its independent parameter space, complicating the training process and inflating computational costs. In response to these challenges, we propose the uni-path learning framework, a more streamlined method that consolidates all AS-ES data into a single model. The training set for the Uni-path Segment generation Model (USM) is thus a combined dataset:

\begin{equation}
    tr(USM)=\{D_{AS},D_{ES}\}
\end{equation}

\begin{algorithm}
\caption{Uni-Path Generation Process}\label{alg:uni_path}
\begin{algorithmic}[1]
\Require USM \Comment{The model gained from uni-path learning}
\State input $\gets$ Start token
\While{True}
    \State USM\_output $\gets$ USM(input)
    \State input $\gets$ input + USM\_output
    \If{stop\_sign \textbf{in} USM\_output}
        \State output $\gets$ input
        \Return output
    \EndIf
\EndWhile
\end{algorithmic}
\end{algorithm}

As detailed in Algorithm \ref{alg:uni_path}, USM undertakes the identical iterative generation procedure as DSM in Algorithm \ref{alg:dual_path}, except that the singular model assumes the responsibilities of both ASM and ESM. This unified approach posits that a single model, given an effective training strategy, can successfully navigate both extractive and abstractive tasks, thereby simplifying the learning process and reducing the requisite resources for model training.
\section{Experiment}
\begin{table*}[th]
  \centering
    \begin{tabular*}{\textwidth}{c|c|c|c|c|c|c}
    \hline
    \hline
    \multirow{2}[4]{*}[1ex]{Metric} & \multirow{2}[4]{*}[1ex]{Para} & \multicolumn{5}{c}{Model} \\
\cline{3-7}          &       & direct & DSM(ent) & USM(ent) & \multicolumn{1}{c}{DSM(inter)} & USM(inter) \\
    \hline
    \multirow{3}[2]{*}{BLEU} & 77M   & \textbf{31.39} & 29.18 ($\downarrow$7.04\%) & 28.53 ($\downarrow$9.11\%) & 28.27 ($\downarrow$9.94\%) & 28.93 ($\downarrow$7.84\%) \\
          & 250M  & \textbf{34.65 } & 30.87 ($\downarrow$10.91\%) & 31.12 ($\downarrow$10.19\%) & 30.63 ($\downarrow$11.60\%) & 29.82 ($\downarrow$13.94\%) \\
          & 800M  & \textbf{33.53} & 31.04 ($\downarrow$7.43\%) & 31.21 ($\downarrow$6.92\%) & 30.08 ($\downarrow$10.29\%) & 29.65 ($\downarrow$11.57\%) \\
    \hline
    \multirow{3}[2]{*}{Acc} & 77M   & 15.25 & 13.74 ($\downarrow$9.90\%) & 16.76 ($\uparrow$9.90\%) & 13.19 ($\downarrow$13.51\%) & \textbf{17.58} ($\uparrow$\textbf{15.28}\%) \\
          & 250M  & 19.23  & 20.60 ($\uparrow$ 7.12\%) & 20.74 ($\uparrow$7.85\%) & 20.60 ($\uparrow$\textbf{7.12}\%) & \textbf{21.70} ($\uparrow$12.84\%) \\
          & 800M  & 22.53 & 24.18 ($\uparrow$7.32\%) & \textbf{24.86} ($\uparrow$\textbf{10.34}\%) & 21.29 ($\downarrow$5.50\%) & 21.57 ($\downarrow$4.26\%)\\
    \hline
    \hline
    \end{tabular*}%
    \caption{Results of different models for Math Word Problem. The best scores and improvements are in \textbf{bold}.}
  \label{tab:mwp_main}%
\end{table*}%
\begin{table*}[th]
  \centering
    % \begin{tabular*}{\textwidth}{c|cc|cc|cc}
    \begin{tabularx}{\textwidth}{>{\centering\arraybackslash}p{1.7cm}|>{\centering\arraybackslash}p{1cm}|>{\centering\arraybackslash}X|>{\centering\arraybackslash}X|>{\centering\arraybackslash}X|>{\centering\arraybackslash}X}
    \hline
    \hline
    \multirow{2}[4]{*}[1ex]{Metric} & \multicolumn{5}{c}{Model} \\
\cline{2-6}          & \multicolumn{1}{c|}{direct} & \multicolumn{1}{c|}{DSM(ent)} & \multicolumn{1}{c|}{USM(ent)} & \multicolumn{1}{c|}{DSM(inter)} & USM(inter) \\
    \hline
    BLEU  & 2.17  & 3.29 ($\uparrow$34.04\%) & \textbf{3.83} ($\uparrow$50.45\%) & 0.37 ($\downarrow$54.71\%) & 0.67 ($\downarrow$45.49\%) \\
    ROUGE-L & 17.15 & 22.25 ($\uparrow$22.92\%) & \textbf{23.96} ($\uparrow$30.60\%) & 7.39 ($\downarrow$43.86\%) & 8.38 ($\downarrow$39.41\%) \\
    MR    & 39.34 & 28.4 ($\downarrow$38.52\%) & \textbf{24.78} ($\downarrow$51.26\%) & 58.87 ($\uparrow$68.76\%) & 33.04 ($\downarrow$22.18\%) \\
    \hline
    \hline
    \end{tabularx}%
    % \end{tabular*}%
    \caption{Results of different models for PET-scan ummarization. The best scores are in \textbf{bold}. Lower MR metric indicates better performance.}
  \label{tab:petscan_main}%
\end{table*}%

\subsection{Dataset}

\paragraph{Math Word Problems (MWP)} For the MWP task, we employ the dataset curated by \citet{fu2023specializing}, which consists of chain-of-thought data generated by the code-davinci-002 model from OpenAI. The original questions are sourced from the GSM8K dataset. \citeauthor{fu2023specializing} enhanced the dataset by prepending four in-context examples to each question to serve as prompts for the GPT model. In our study, we extract only the original questions and their corresponding answers augmented with CoT as our dataset.

\paragraph{PET Report Summarization (PET)}For the PET task, we introduce the cPET-11K dataset, a novel compilation of 11.6k Chinese PET/CT report-impression pairs. This dataset is a collection of PET/CT report data from patients with pancreatic cancer, originating from real clinical data of a major tertiary hospital. These PET/CT scans were performed using three PET/CT machines. All patient reports have been anonymized, with only the content of the reports retained. These reports focus on determining whether the patient has pancreatic cancer and whether there is distant metastasis of the pancreatic cancer to other abdominal organs.

\subsection{Implementation}

\paragraph{Base Model} We use two variants of the typical seq2seq model T5 \cite{2020t5}. For MWP, we follow \citet{fu2023specializing} and use Flan-T5 \cite{chung2022ft5}. For PET impression generation, we use mt5 \cite{xue2020mt5} for its multilingual capabilities. We conduct experiments with three different sizes of Flan-T5 (small, base, large) to investigate the impact of model size on the efficacy of AS-ES learning. Unless otherwise specified, the default models used are Flan-T5-base for MWP and mT5-base for PET impression generation.

\paragraph{Training Process} As the size of the AS-ES dataset is usually bigger than the original dataset, to ensure a fair comparison among different methods, all uni-Path approaches are trained with the same amount of batch size and learning rate given the same amount of training time. Conversely, dual-path approaches undergo training for around half the duration per model in accordance with their data amount.

\paragraph{Checkpoint Selection} Considering the final generated results can not be directly obtained during evaluation for iterative approaches like AS-ES learning, traditional evaluation metrics like BLEU score or validation loss may not directly correlate with actual model performance. Furthermore, even for a direct approach, a higher BLEU score or lower loss does not necessarily lead to a higher accuracy for MWP. As for PET scan impression generation, the BLEU score can be significantly influenced by variations in formatting, necessitating additional post-processing steps that can alter the metric's relevance to actual performance.

To address the above issues, we select three different checkpoints for each approach, and report the best performance among the three. The detailed criteria are as follows. (1) best\_train: the model with the lowest loss on the training set. (2) best\_loss: the model with the lowest loss on the validation set. (3) best\_bleu: the model with the highest BLEU on the validation set. For MWP, accuracy is deemed the primary performance indicator, while for PET, we prioritize BLEU scores.

\begin{equation}
\label{mr}
    MR=\frac{1}{N} \sum_{i=1}^{N} \left( \frac{\left| R_{\text{GT}_i} - (R_{\text{GT}_i} \cap R_{\text{GR}_i}) \right|}{\left| R_{\text{GT}_i} \right|} \right)
\end{equation}

\paragraph{Evaluation Metrics} For MWP, we use BLEU \cite{papineni2002bleu} and accuracy as the evaluation metrics with a focus on accuracy. For PET, we use BLEU, ROUGE \cite{lin2004rouge} and MR (missing ratio) as the evaluation metrics with a focus on BLEU. MR is calculated using Equation \ref{mr}, assessing the proportion of anomalies omitted in the generated summaries. Here, 
$R$ denotes the various organs or regions identified via keyword mapping, $GT$ and $GR$ represent the ground truth and generated results respectively, and $N$ is the size of the test set.

\subsection{Overall Performance of AS-ES Learning}

As AS-ES learning has multiple combinations of segmentation strategy and training strategy, we here report the most effective segmentation for the two tasks respectively, and apply both USM and DSM for the two segmentation strategies. As shown in Table \ref{tab:mwp_main} and Table \ref{tab:petscan_main}, AS-ES learning improves the model performance on both tasks with the appropriate strategy. Entropy-oriented segmentation shows a generalizability across different model sizes and tasks, while interleaving segmentation is more targeted.

We here summarize some key aspects in the experiments as a summary of the characteristics of AS-ES learning, as well as an attempt to answer the question mentioned in Section \ref{intro}, that is, whether the inferiority of small models in CoT-related tasks stem from its inherent incapability to do extraction and reasoning with a singular model.

\subsubsection{Lower BLEU with Higher Accuracy}

The first thing to notice in Table \ref{tab:mwp_main} is that, although all AS-ES learning leads to lower BLEU score compared to the direct approach, most of them leads to a higher accuracy. This may implicate that instead of memorizing the solution provided in the training set in direct approach, AS-ES learning 
enables logical reasoning of higher granularity in small models through a literal step-by-step process. Models trained through AS-ES learning could yield different solution and therefore although differs from the ground truth in terms of textual similarity, still leads to the same correct results.

\subsubsection{Different Model Size Reacts Differently to Segmentation Strategy}

As shown in Table \ref{tab:mwp_main}, entropy-oriented segmentation works best for Flan-T5-large while interleaving segmentation works best for Flan-T5-small. The reasons behind this could be two-fold. First, entropy calculation is the key for entropy-oriented segmentation. The inherent superiority of larger models after fine-tuning results in a better grasp of data, leading to a better segmentation.
Furthermore, when it comes to task of highly sequential CoT solution like MWP, entropy-oriented segmentation does not seem that straightforward and easy for small models to grasp, especially with the suboptimal entropy calculation by small models. Overall, the specific performance of AS-ES learning depends on both the quality of AS-ES dataset and the capacity of models to handle to data.

% Plus, when using interleaving segmentation, AS-ES dataset inevitably gains more data, resulting in longer training time compared to entropy segmentation. Therefore, for bigger models with enough capacity in complexity handling, given the same amount of training time, it performs better with entropy segmentation but worse with interleaving segmentation.

\subsubsection{Segmentation Works Differently for Different Tasks}

As shown in Table \ref{tab:mwp_main} and Table \ref{tab:petscan_main}, PET scan summarization and MWP show rather large divergence using entropy segmentation and interleaving segmentation. The former benefits largely from entropy segmentation while suffers a lot from interleaving segmentation. On the contrary, interleaving segmentation works for most cases in MWP, depending on the size of the model. The reason behind this is quite intrinsic. The structure of math word problems typically follows a logical progression where statements provide context or premises (ES) followed by a step in reasoning or calculation (AS), and this pattern tends to repeat as the problem is broken down into solvable parts therefore naturally aligns with the interleaving segmentation approach. Impression generation from PET scans, on the other hand, tend to consist of multiple observations (ES) followed by a collective diagnostic insight (AS), or vice versa without an interleaving pattern, therefore better accommodates to entropy segmentation. 

% \subsubsection{Synergy of segmentation strategy and model size}

\subsubsection{USM v.s. DSM: Is One Model Enough?}

Although existing works all use two separate models for iterative generation approach, experiments results for both MWP and PET suggest that one model is enough and lead to even better results compared to using two models, at least for AS-ES learning. Although both suffering from error accumulation, USM consistently outperforms direct approach compared to DSM. The integrated context for training USM might enable the model to better understand the interplay between different types of reasoning, leading to a more nuanced understanding of the data. Furthermore, the separate model in DSM leads to separate optimization, making the checkpoint used for test may not be fully display its capacity, even with multiple checkpoint selection strategy. This phenomenon suggests that it is more about the training strategy but rather the capacity of model itself.

% One interesting anomaly is that, flan-t5-small is the only one variant that sees an increase with USM but a decrease with DSM under the same segmentation strategy. 
%  

\section{Discussion}
\subsection{Effect of Segmentation}
\begin{table}[th]
  \centering
    \begin{tabular}{c|cc}
    \hline
    \hline
    Segment & BLEU  & Accuracy \\
    \hline
    Baseline & \textbf{34.65} & 19.23 \\
    Inter & 30.62 ($\downarrow$11.63\%) & \textbf{20.60} ($\uparrow$7.12\%) \\
    ent   & 30.87 ($\downarrow$10.90\%) & \textbf{20.60} ($\uparrow$7.12\%) \\
    ent*  & 30.88 ($\downarrow$10.88\%) & 18.96 ($\downarrow$1.40\%) \\
    BLEU  & 29.97 ($\downarrow$13.50\%) & 17.72 ($\downarrow$7.85\%) \\
    ROUGE & 27.46 ($\downarrow$20.75\%) & 16.21 ($\downarrow$15.70\%) \\
    Loss  & 30.34 ($\downarrow$12.43\%) & 20.19 ($\uparrow$4.99\%) \\
    \hline
    \hline
    \end{tabular}%
    \caption{Results of DSM for MWP using different segmentation method. Baseline refers to direct approach. The best scores are in \textbf{bold}.
    % Inter refers to location-oriented segmentation strategy. Ent refers to entropy-oriented segmentation using model trained through direct approach, while ent* uses pre-trained but not fine-tuned model for entropy calculation. BLEU and Rouge are the two derivatives of similarity-oriented segmentation, while Loss denotes the loss-oriented segmentation.
    }
  \label{tab:segment}%
\end{table}%

% \begin{figure}[t]
%   \centering
%   \includegraphics[width=\linewidth]{figure/ratio.pdf}
%   \caption{
%   Results of ablation of ratio.
%   }
%   \label{fig:ratio}
% \end{figure}

% \begin{figure}[t]
%   \centering
%   \includegraphics[width=\linewidth]{figure/beta.pdf}
%   \caption{
%   Results of ablation of beta.
%   }
%   \label{fig:beta}
% \end{figure}

% \begin{figure}[t]
%   \centering
%   \includegraphics[width=\linewidth]{figure/gamma.pdf}
%   \caption{
%   Results of ablation of gamma.
%   }
%   \label{fig:gamma}
% \end{figure}

\begin{figure*}[ht]
  \centering
  \begin{subfigure}[b]{0.3\textwidth} % Width is set to 30% of text width
    \includegraphics[width=\textwidth]{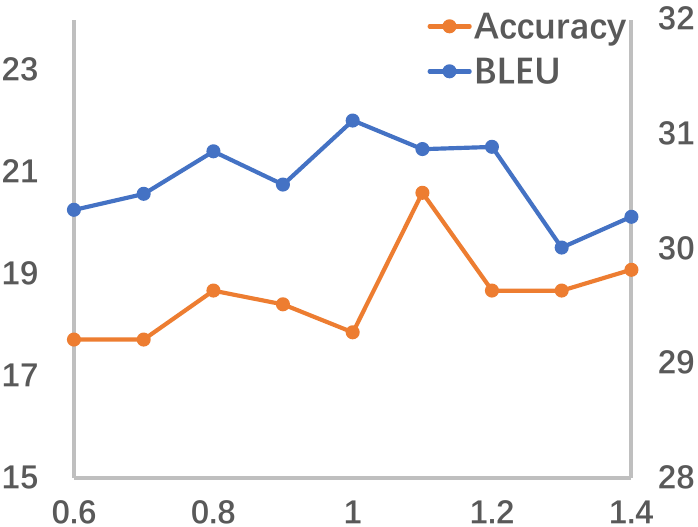}
    \caption{DSM for MWP with different $\beta$}
    \label{fig:hyp1}
  \end{subfigure}
  \hfill % This will insert a space between the two figures
  \begin{subfigure}[b]{0.3\textwidth}
    \includegraphics[width=\textwidth]{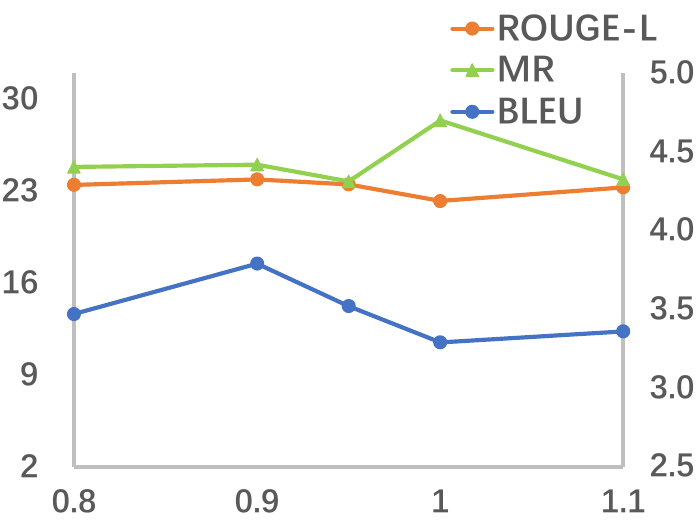}
    \caption{USM for PET with different $\beta$}
    \label{fig:hyp2}
  \end{subfigure}
  \hfill
  \begin{subfigure}[b]{0.3\textwidth}
    \includegraphics[width=\textwidth]{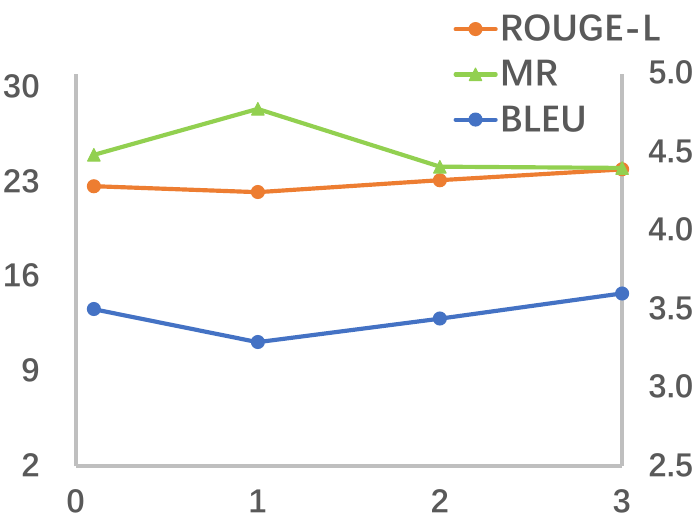}
    \caption{USM for PET with different $\gamma$.}
    \label{fig:hyp3}
  \end{subfigure}
  \caption{results of different hyperparameter settings for AS-ES learning. The BLUE metric values (blue line) correspond to the right Y-axis (secondary axis). $\gamma$ is set to $1.0$ in Figure \ref{fig:hyp2}, while $\beta$ is set to $1.0$ in Figure \ref{fig:hyp3}.}
  \label{fig:images}
\end{figure*}

\begin{figure*}[ht]
  \centering
  \begin{subfigure}[b]{0.24\textwidth} % Width is set to 30% of text width
    \includegraphics[width=\textwidth]{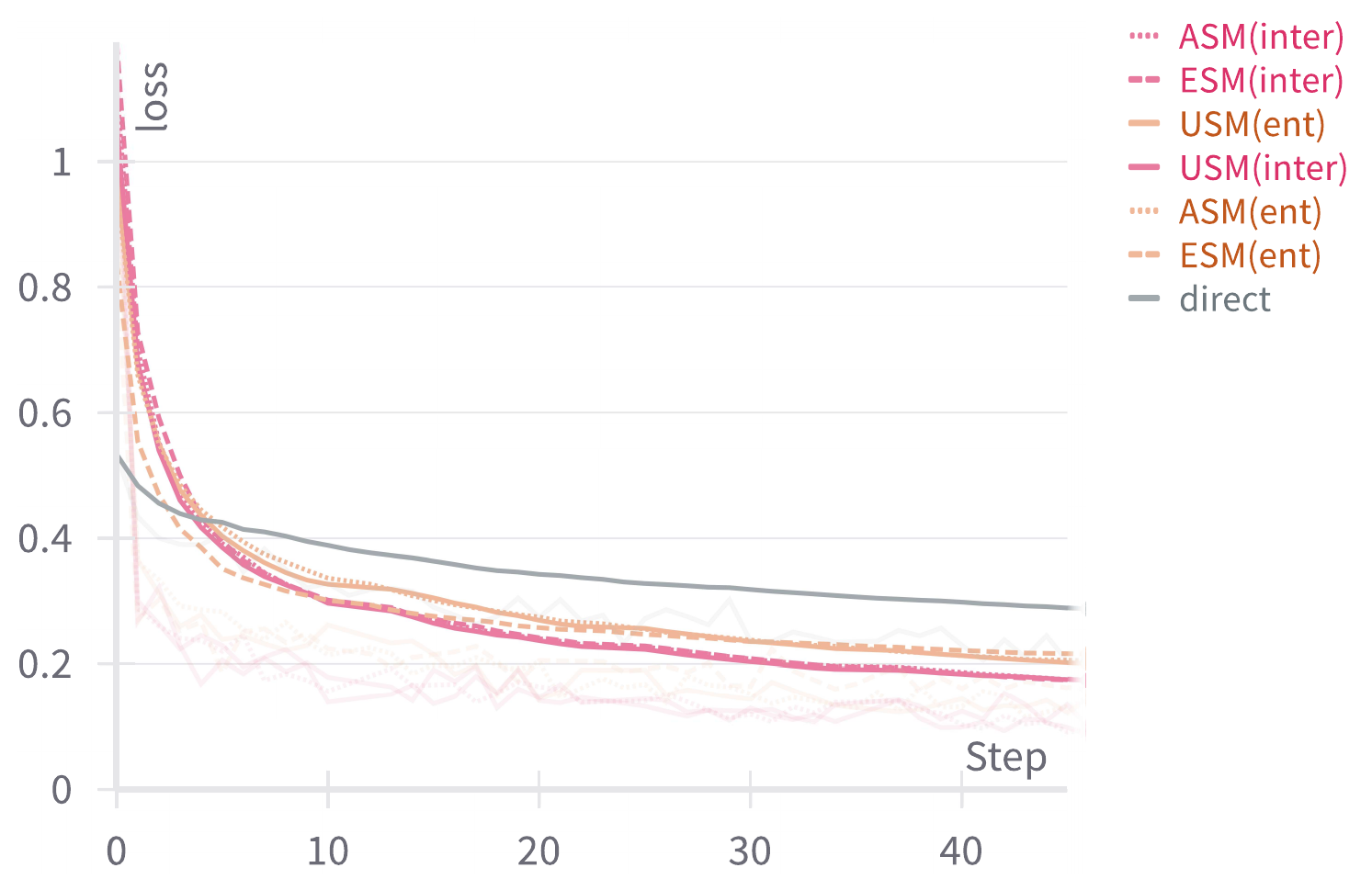}
    \caption{AS-ES learning with entropy segmentation for MWP}
    \label{fig:dis1}
  \end{subfigure}
  \hfill % This will insert a space between the two figures
  \begin{subfigure}[b]{0.24\textwidth}
    \includegraphics[width=\textwidth]{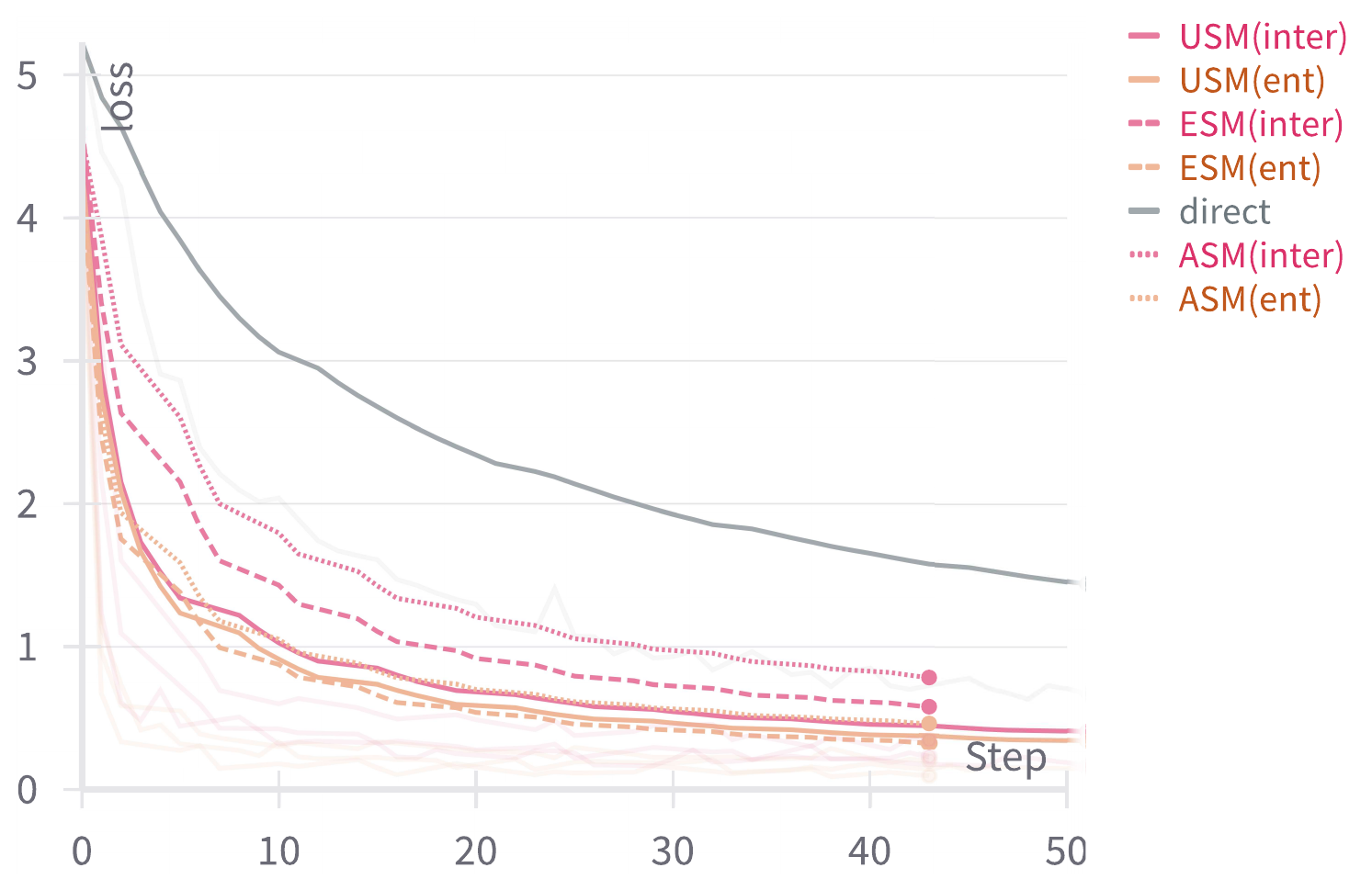}
    \caption{AS-ES learning with entropy segmentation for PET}
    \label{fig:dis2}
  \end{subfigure}
  \hfill
  \begin{subfigure}[b]{0.24\textwidth}
    \includegraphics[width=\textwidth]{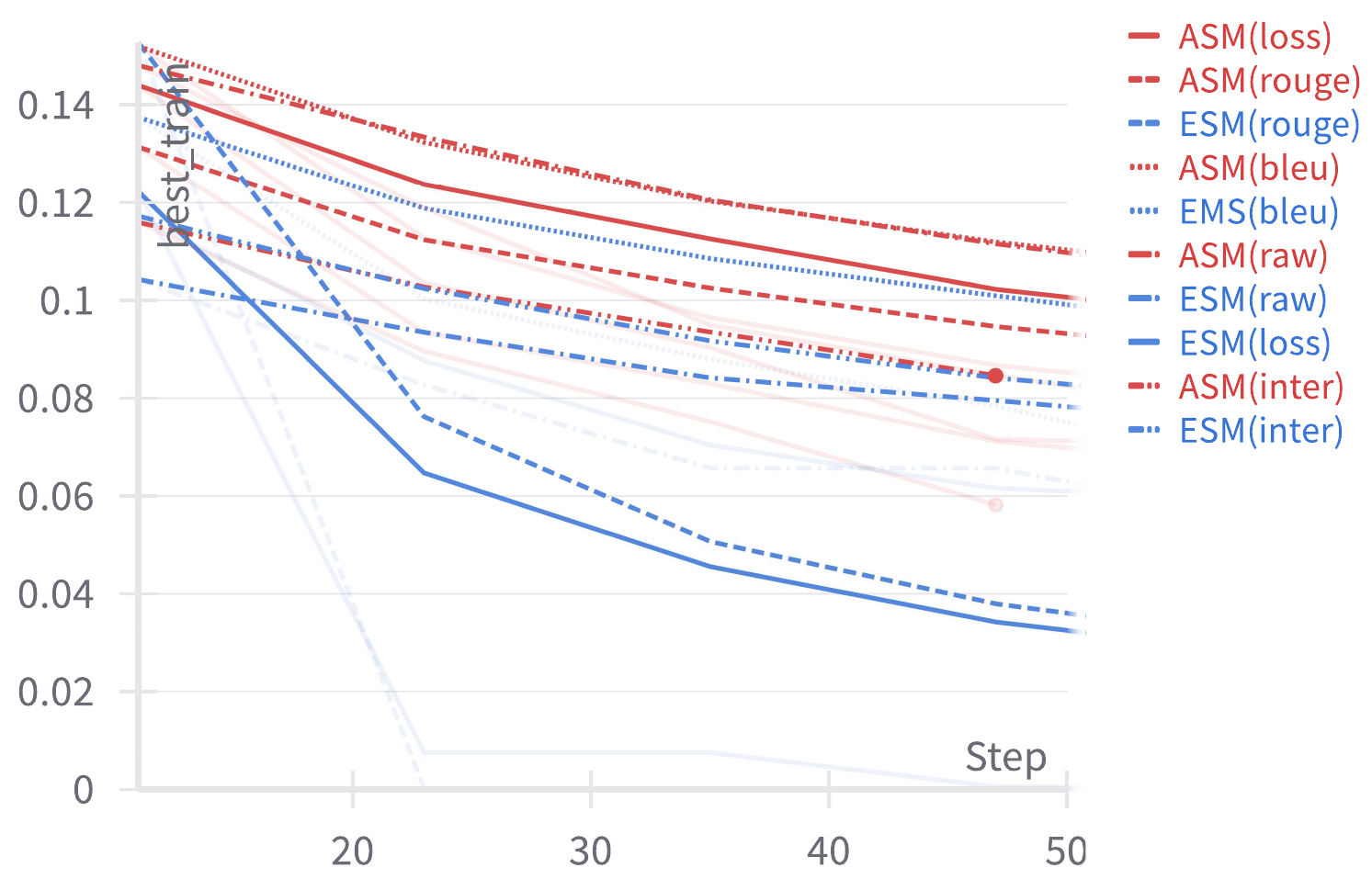}
    \caption{ASM/ESM with different segmentation for MWP}
    \label{fig:dis3}
  \end{subfigure}
  \hfill % This will insert a space between the two figures
  \begin{subfigure}[b]{0.24\textwidth}
    \includegraphics[width=\textwidth]{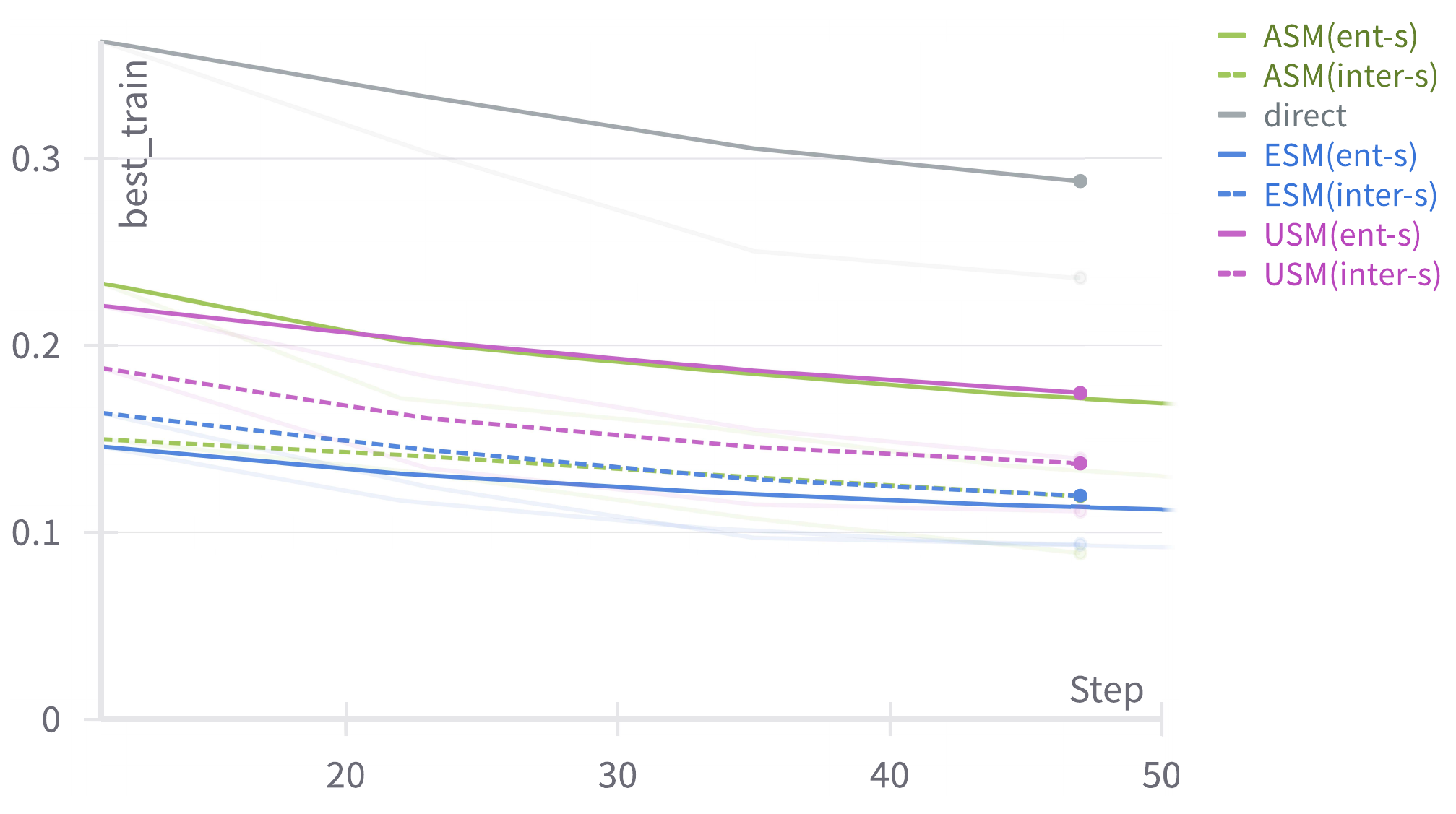}
    \caption{USM/ASM/ESM with entropy segmentation for MWP}
    \label{fig:dis4}
  \end{subfigure}
  \caption{Smoothed training curve for AS-ES Learning. In Figure \ref{fig:dis1} and Figure \ref{fig:dis2}, 
  % show that the performance segmentation strategy varies among tasks. 
  Grey, pink and yellow denote direct approach, interleaving segmentation and entropy segmentation respectively. Figure \ref{fig:dis3} and Figure \ref{fig:dis4} show the curve of the best training loss among time. All curves for ASM is displayed in red with ESM in bleu in Figure \ref{fig:dis3}. In Figure \ref{fig:dis4}, curves for ASM, ESM, USM are displayed in green, blue and pink, respectively.}
  \label{fig:images}
\end{figure*}
As shown in Table \ref{tab:segment}, different segmentation strategies do play a key role in AS-ES learning. segmentation by interleaving, loss and entropy calculated from fine-tuned models all yield better results compared to the direct approach. The decreased performance using BLEU/ROUGE segmentation indicates that simply segmenting AS/ES by its textual similarity (as what is intrinsic to humans) is not the same way for models. Furthermore, using entropy calculated by pre-trained but not fine-tuned models introduces no further improvement, which makes sense since pre-trained models have less grasp about which part is extractive/abstractive.

\subsection{Effect of Hyperparameters}

Most segmentation methods introduce a hyperparameter $\beta$ as in Equation \ref{as-equation} and \ref{es-equation}, and PET summarization introduces a hyperparameter of $\gamma$ which evaluates the amount of incorporated normality. Here we evaluate how these hyperparameters affect different training strategies for different tasks.

\paragraph{Effect of Ratio $\beta$}

As shown in Figure~\ref{fig:hyp1} and Figure~\ref{fig:hyp2}, entropy segmentation for both USM and DSM for different tasks has a $\beta$ threshold (around 1) where the model achieves the best balance between ES and AS.

\paragraph{Effect of Ratio $\gamma$} 

As shown in Figure~\ref{fig:hyp3}, $\gamma$ at $1.0$ reaches the lowest point on the curve, which might be a tipping point where the inclusion of normal findings is enough to dilute the model's focus on anomalies without providing the additional contextual benefits seen at higher $\gamma$ values. Overall speaking, the model performance is less affected by $\gamma$ compared to segmentation and training strategy.

\subsection{Why AS-ES Learning Works?}
\label{dis:mechanism}
We here further explore the underlying mechanism of AS-ES learning to see why it works. As shown in Figure~\ref{fig:dis3}, the best training loss boundary of ESM is generally lower than ASM, which aligns with our hypothesis that extraction comes more easily while logical reasoning is not quite so. Figure~\ref{fig:dis4} shows the best training loss boundary for the direct approach, USM, ASM and ESM respectively. As expected, the lowest loss boundary for the direct approach is significantly larger than AS-ES learning, which partly explains why AS-ES learning works. The divergence of MWP and PET to segmentation strategy could also be explained from this perspective. As shown in Figure~\ref{fig:dis1} and \ref{fig:dis2}, interleaving segmentation generally reaches a lower boundary than entropy segmentation for MWP therefore resulting in a better performance using interleaving segmentation, and vice versa for PET. The overall discovery suggests that AS-ES learning works by achieving a generally lower loss boundary compared to the direct approach.
% \subsection{}
% \subsection{}
% \subsection{The underlying mechanism of AS-ES learning}
% \begin{itemize}
%     \item The balance between data : effect of ratio
%     \item relax the model : does it work with only one model?
%     \item lower loss boundary
% \end{itemize}
\section{Conclusion}

In this paper, we introduce a data-efficient CoT distillation strategy for small models. By segmenting CoT data into extractive part and abstractive part respectively, we improve the model performance through an iterative generation approach without incorporating additional data. The generalization of AS-ES learning to different model sizes and tasks shows its effectiveness. We further explore whether two models are necessary for this interactive generation approach, and answer the derivative question that the limitation of small models in CoT learning stems from the training paradigm instead of its inherent capacity, providing insights into the underlying mechanism of CoT.
\section*{Limitations}

In this paper, we mainly experiment and discuss the use of AS-ES Learning based on the direct training approach. Although experiments show that the use of one single model is better than two separate models, this may not be the case when there is a solution for DSM to be simultaneously trained. Furthermore, although we come up with a general range of the hyperparameter settings for AS-ES learning, the specific optimal settings of AS-ES Learning on different datasets may vary and therefore require a need for specific exploration.
\section*{Ethical Considerations}
In this work, we introduce a new Chinese PET report-impression dataset. The data collection protocol is approved by an ethics review board.  All experimental datasets involved have been de-identified by dataset providers and used for research only.
\bibliography{ref}

\appendix

\section{Details of Implementation}
\label{sec:appendix}

Our model utilizes the Pytorch-based \citep{paszke2019pytorch} Huggingface Transformers \citep{wolf2020transformers} packages. All experiments are conducted with the same batch size of 64 and the same learning rate of 5e-4 for MWP 1e-3 for PET using early stopping. Both MWP and PET datasets are split into \emph{train}, \emph{validation}, \emph{test} sets with a ratio of 80\%, 10\%, and 10\% respectively. The results reported are the average of three separate runs. Most experiments were conducted on NVIDIA A100-80GB-PCIe GPUs or A100-SXM4-80GB GPUs, some were conducted on Tesla V100S-PCIE-32GB GPUs. Code and dataset will be released upon acceptance.
\end{document}